\documentclass[10pt,twocolumn,letterpaper]{article}

\usepackage{cvpr}
\usepackage{times}
\usepackage{epsfig}
\usepackage{graphicx}
\usepackage{amsmath}
\usepackage{amssymb}
\usepackage{algorithm}
\usepackage[noend]{algpseudocode}
\usepackage{booktabs}

\cvprfinalcopy 

\begin{document}

\title{Accelerating Large Scale Knowledge Distillation via \\Dynamic Importance Sampling}

\author{Minghan Li$^1$\thanks{The authors contribute equally to this paper.} , Tanli Zuo$^{1 *}$, Ruicheng Li$^1$, Martha White$^{2 \dagger}$, Weishi Zheng$^1$\thanks{Corresponding Authors.}\\
$^1$School of Data and Computer Science, Sun Yat-Sen University \\
$^2$Department of Computing Science, University of Alberta\\
$\left \{ \text{limh23, zuotl, lirch5} \right \}\text{@mail2.sysu.edu.cn}$ \\ whitem@ualberta.ca \  wszheng@ieee.org}
\date{}

\maketitle

\begin{abstract}
   Knowledge distillation is an effective technique that transfers knowledge from a large teacher model to a shallow student. However, just like massive classification, large scale knowledge distillation also imposes heavy computational costs on training models of deep neural networks, as the softmax activations at the last layer involve computing probabilities over numerous classes. In this work, we apply the idea of importance sampling which is often used in Neural Machine Translation on large scale knowledge distillation. We present a method called dynamic importance sampling, where ranked classes are sampled from a dynamic distribution derived from the interaction between the teacher and student in full distillation. We highlight the utility of our proposal prior which helps the student capture the main information in the loss function. Our approach manages to reduce the computational cost at training time while maintaining the competitive performance on CIFAR-100 and Market-1501 person re-identification datasets.
\end{abstract}

\section{Introduction}
In the last few years, deep neural networks have achieved state-of-the-art performance in applications like image recognition \cite{he2016deep,szegedy2015going} and natural language processing \cite{jozefowicz2016exploring,wu2016google}, yet large scale classification such as face verification \cite{liu2016large,schroff2015facenet,wen2016discriminative} and neural machine translation \cite{jean2014using,morin2005hierarchical} still remains challenging. The main difficulty of such massive classification tasks comes from the last softmax layer of modern deep neural nets. Computing the full activations often involves calculating probabilities over all classes in the normalization constant, which requires substantial  computational power to compute its dot product with the last hidden layer of the neural network.

\begin{figure}[t]
\centering 
\includegraphics[scale=0.3]{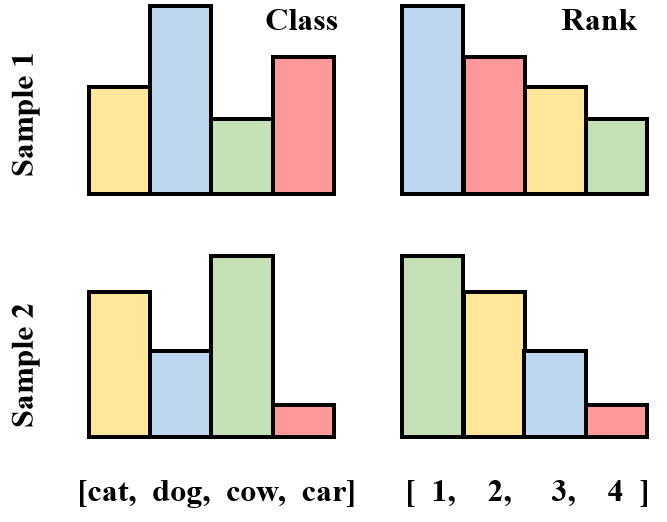}
\caption{Predictions over classes and ranks of two different samples. We sort the classes in descending order of the teacher model's prediction to obtain the class ranking. The term ``rank'' will be frequently used in this work.}
\label{fig:semantic_sorted}  
\end{figure}

The same problem also occurs in large scale knowledge distillation. Knowledge distillation is a model compression technique where a shallow student model tries to mimic the output of a large complex teacher model. Similar to the regular training paradigm, knowledge distillation also suffers from a growing time complexity of computing softmax probabilities over a large number of classes. To mitigate the problem, the specialists ensemble \cite{hinton2015distilling} was proposed where each specialist was assigned to a subset of data and learned from a single teacher in parallel. Co-distillation \cite{anil2018large} trained multiple neural nets together on disjoint sets of data and encouraged them to share knowledge with each other. Although these methods managed to accelerate training by a large margin, more parameters are involved since we are training multiple models in parallel, which needs massive computational power and storage space, making it difficult to deploy on mobile devices.

Aiming to efficiently distill the knowledge from a teacher model on large scale datasets, we apply the sampling based approaches \cite{bengio2008adaptive,jean2014using} which is often used in Neural Machine Translation. However, such methods often require a prior distribution of the word frequency (such as a unigram), making it difficult to extend to other areas. Fortunately, in knowledge distillation \cite{hinton2015distilling} it is very likely that we can obtain this prior from an \textit{oracle} that already generalizes well, yet we find scant amount of research in this direction.

\begin{figure}[t]
\centering 
\includegraphics[scale=0.33]{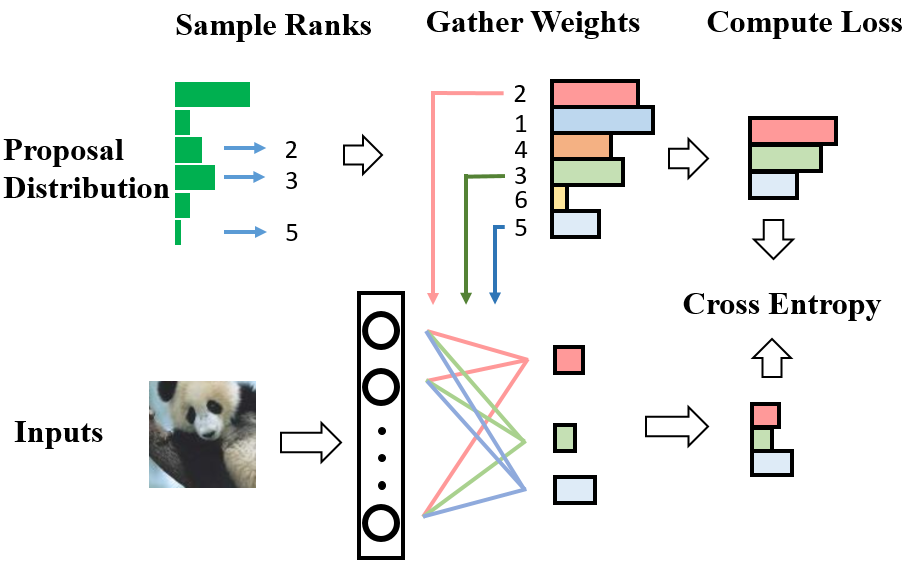}
\caption{Forward-view of importance-sampling based distillation. We first relabel the dataset with the teacher's prediction, i.e. the soft labels. We then sample ranks from the proposal distribution and find the corresponding subset of classes. We finally compute the cross-entropy loss between the teacher's and student's predictions over the subset. This foward-view is for illustration. For implementation, we use the backward-view described in  Alg. \ref{alg:sampling_distillation}.}
\label{fig:ours}  
\end{figure}

In this work, we present a simple yet effective approach called \textit{dynamic importance sampling}, which samples a subset of ranked classes from a proposal distribution that is dynamically adjusted during training. The proposal prior is derived from a method called  \textit{dynamic class selection} \cite{zhang2018accelerated}, which enables the student to back-propagate the main information in the loss function without computing the full softmax activation. We also compare our method with fixed importance sampling approaches which sample from a uniform distribution or directly from the teacher's prediction. We show that neither of these two methods perform well when the sampled subset is small compared to using the dynamic distribution. Our approach reduces computational costs during training and sometimes even outperforms the full distillation on CIFAR-100 and Market-1501 person re-identification datasets. 

\section{Background} 
Suppose we have a training set consisting of pairs of sample and label $(x, y)$, where $y \in C\text{: }C$ is the set of all classes. In order to avoid confusion when deriving the importance-sampling based distillation in the following section, we adopt the terms in energy based models \cite{teh2003energy} to describe the basic components of a neural net. Given the input and network's parameters, the energy function of the neural network is:

\begin{equation}
    \mathcal{E} = -\Phi(x;\theta)W
\end{equation}
where $\Phi(\cdot;\theta)$ is the final representation, $W$ is the weight matrix at the last layer and $\theta$ denotes all the trainable parameters in the network. To obtain the prediction of the student network, we need to normalize the exponential energy:

\begin{equation}
    q = \frac{e^{-\mathcal{E}}/T}{Z} \text{ , with } Z =\sum_{i} e^{-\mathcal{E}_i/T}
\end{equation}
where $T$, which is called the temperature parameter, controls the entropy of the network's prediction $q$. As $T \rightarrow \infty$, $q$ gradually converges to the uniform distribution. In practice, we often pick a medium temperature so as to reveal sufficient inter-class information in the teacher's prediction $p$. Let $q$ be the prediction of the student. We can present our loss function as: 
\begin{equation} \label{eq:distillation}
  \mathcal{L}(q, y, p)= \lambda\ell (q, p)+(1-\lambda) \ell (q, y)  
\end{equation}
where $\ell$ is the cross entropy loss and $\lambda$ is a hyperparameter that balances the two cross entropy losses. \cite{romero2014fitnets} shows that distillation can be seen as a special form of curriculum learning if $\lambda$ is gradually decreased as training proceeds. However, in this paper we set $\lambda=1$ in all our experiments for simplicity and clarity.


\section{Methodology}
In this section, we formalize our approach in an importance-sampling based framework \cite{bengio2008adaptive,jean2014using} which samples classes from a proposal distribution. We derive a mixture of Laplace distributions whose the parameters can be dynamically adjusted during training from the dynamic class selection process\cite{zhang2018accelerated}. Our method not only speeds up the training significantly but also maintain a competitive performance to the full distillation. 

\subsection{Importance-Sampling Based Distillation}
The main idea of importance-sampling based distillation is to approximate the expected gradients of the full energy function with the one computed over a set of sampled classes. Moreover, instead of directly sampling from the student's prediction $q$ which is costly to compute, we sample from a proposed prior distribution $r$ to estimate the expected gradients. We formalize our approach using the framework of importance-sampling based approximation \cite{bengio2008adaptive,jean2014using} which avoids computing the full matrix multiplication at the softmax layer. The gradients of the cross-entropy loss in Equation (\ref{eq:distillation}) w.r.t the model's parameters $\theta$ over the complete set of classes $C$ are:

\begin{equation} \label{eq:energy_grad}
    \begin{aligned}
  \nabla_{\theta}\mathcal{L} &= \sum_{i\in C} (p_i-q_i)\nabla_{\theta}\mathcal{E}^s_i\\
   &= \sum_{i\in C}p_i\nabla_{\theta}\mathcal{E}^s_i - \sum_{i\in C}q_i\nabla_{\theta}\mathcal{E}^s_i \\
   &= E_p[\nabla_{\theta}\mathcal{E}^s] - E_q[\nabla_{\theta}\mathcal{E}^s] \\
   \end{aligned}
\end{equation}
where $p$ is the teacher's prediction, $q$ is the student's prediction and $\mathcal{E}^s$ is the energy function of the student. The main difficulty here is to estimate both $E_p[\nabla_{\theta}\mathcal{E}^s]$ and $E_q[\nabla_{\theta}\mathcal{E}^s]$ when the number of classes is large. Therefore, we need to sample from another pre-defined distribution to efficiently estimate the expectation. If we have a proposal distribution $r$ such as the one in Fig \ref{fig:ours}, we can approximate this expectation by importance sampling:

\begin{equation}
    \begin{aligned}
    E_q[\nabla_{\theta}\mathcal{E}^s] &= \sum_{i\in C}q_i\nabla_{\theta}\mathcal{E}^s_i\\
    &=\sum_{i\in C}r_i \cdot \frac{q_i}{r_i}\nabla_{\theta}\mathcal{E}^s_i\\
    &= E_r\left [\frac{q_i}{r_i}\nabla_{\theta}\mathcal{E}^s_i \right ].\\
    \end{aligned}
\end{equation}

\begin{algorithm}[t]
\caption{Backward-View of Important Sampling Based Distillation}
\label{alg:sampling_distillation}
\begin{algorithmic}[]
    \State $V \gets 0\text{ , } g \gets 0$ \text{ // Initialization}
    \vspace{1mm}
    \State $U \gets 0\text{ , } h \gets 0$
    \vspace{1mm}
    \State $\nabla_{\theta}\mathcal{L} \gets 0$
    \vspace{1mm}
    \State $v \gets e^{-\mathcal{E}^s_i}\text{ , }u \gets e^{-\mathcal{E}^t_i}$\text{ // Add target class}
    \vspace{1mm}
    \State $g \gets g + v \nabla_{\theta}\mathcal{E}^s_i\text{ , }h \gets h + u \nabla_{\theta}\mathcal{E}^t_i$
    \vspace{1mm}
    \State $V \gets V + v\text{ , }U \gets U + u$
    \vspace{1mm}
    \For{j=1 \textbf{to} m}
        \vspace{1mm}
        \State $k \sim r(\cdot)$ \text{ // Sample negative classes}
        \vspace{1mm}
        \State $v \gets \frac{e^{-\mathcal{E}^s_k}}{r_k}\text{ , }u \gets \frac{e^{-\mathcal{E}^t_k}}{r_k}$
        \vspace{1mm}
        \State $g \gets g + v \nabla_{\theta}\mathcal{E}^s_k\text{ , }h \gets h + u \nabla_{\theta}\mathcal{E}^s_k$
        \vspace{1mm}
        \State $V \gets V + v\text{ , }U \gets U + u$
    \EndFor
    \vspace{1mm}
    \State $\nabla_{\theta}\mathcal{L} \gets \nabla_{\theta}\mathcal{L}+\frac{1}{U}h - \frac{1}{V}g$
\end{algorithmic}
\end{algorithm}

However, although we don't have to sample from $q$ anymore, we still need to compute $q$ over all the classes. \cite{bengio2008adaptive} proposed a biased but more efficient version of importance sampling estimator to $E_q[\nabla_{\theta}\mathcal{E}^s]$:
\begin{equation}
\begin{aligned}
    &\frac{1}{V}\sum_{i\in S} v_i\nabla_{\theta}\mathcal{E}^s_i
    \end{aligned}
\end{equation}
where $V= \sum_{i\in S}v_i $ and $v_i = e^{-\mathcal{E}^s_i}/r_i$. $S$ is a subset of classes sampled from $r$ with replacement. Though this estimator is biased, it was shown that \cite{bengio2008adaptive}  the estimation converges to the true mean as $\left | S \right | \to \infty$. So the gradients in Equation (\ref{eq:energy_grad}) can be approximated by:
\begin{equation}
    \begin{aligned}
    \frac{1}{U}\sum_{i\in S} u_i\nabla_{\theta}\mathcal{E}^s_i -  \frac{1}{V}\sum_{i\in S} v_i\nabla_{\theta}\mathcal{E}^s_i
    \end{aligned}
\end{equation}
where $U= \sum_{i\in S}u_i $, $u_i = e^{-\mathcal{E}^t_i}/r_i$ and $\mathcal{E}^t$ is the energy function of the teacher. Since we manually add the target class $i$ to the sampled subset, $r_i$ is set to 1 when computing $u_y$ and $v_y$. We describe the importance-sampling based distillation in  Alg. \ref{alg:sampling_distillation}. As we can see, the proposal distribution plays an important role in our method. The default option for this prior is usually the uniform distribution, which assumes that we have no prior information about the frequency distribution of classes. However, in distillation we can utilize the teacher model to derive our own proposal distribution. Ideally, this prior should tell us about the main information we need to back-propagate.  

\subsection{Prediction-Difference based Selection}
Before we present our design of the dynamic distribution, we need to first introduce a dynamic class selection method \cite{zhang2018accelerated} which we refer to as the prediction-difference based selection (PDBS) in this paper, for it is the key-stone to derive the proposal prior. The basic idea of the prediction-difference based selection in distillation is to select classes that have the largest absolute difference between the teacher's and student's prediction. After the selection stage, we use the selected subset of classes $S$ to approximate the full softmax activation:

\begin{equation} \label{eq:selective_softmax}
  \begin{aligned}
  q_i &= \frac{\exp(-\mathcal{E}^{'}_i/T)}{\sum\limits_m \exp(-\mathcal{E}^{'}_m/T)} \text{ , }\\
  \text{with } \mathcal{E}^{'}_i &= -\widetilde W_i^T \Phi(x;\theta) \text{ , } \forall i \in S
  \end{aligned}
\end{equation}
where $\widetilde W_i \in \mathbb{R}^{\left | S\right |}$ is the submatrix of the complete weight matrix $W \in \mathbb{R}^{d}$ and $\left | S\right | \ll d$. 

The major assumption behind PDBS is that most gradients are concentrated on the classes that have the biggest absolute difference between the predictions and labels. An empirical study \cite{zhang2018accelerated} shows that the gradients w.r.t. logits of the classes that have the k biggest absolute values indeed take up the most proportion. And we know from Equation (\ref{eq:energy_grad}) that the gradients are proportional to the difference between the labels and predictions, which explains why using the prediction difference to select the classes. To avoid confusion, it is worth mentioning that this selection method is deterministic while the sampling approach introduces randomness. We make further comparison between these two approaches and analyze the experiment results of them in a later section. 

Although this method shows a competitive performance to the one trained by full softmax, in order to obtain the k largest absolute prediction difference, it still involves computing a full softmax activation in the student's prediction. In the next section, we use a mixture of distributions to approximate the dynamics of the PDBS method. Combining the importance sampling technique, we are exempt from querying the whole softmax activation of the student model while providing an effective approximation.

\begin{figure}[t]
\centering 
\includegraphics[scale=0.38]{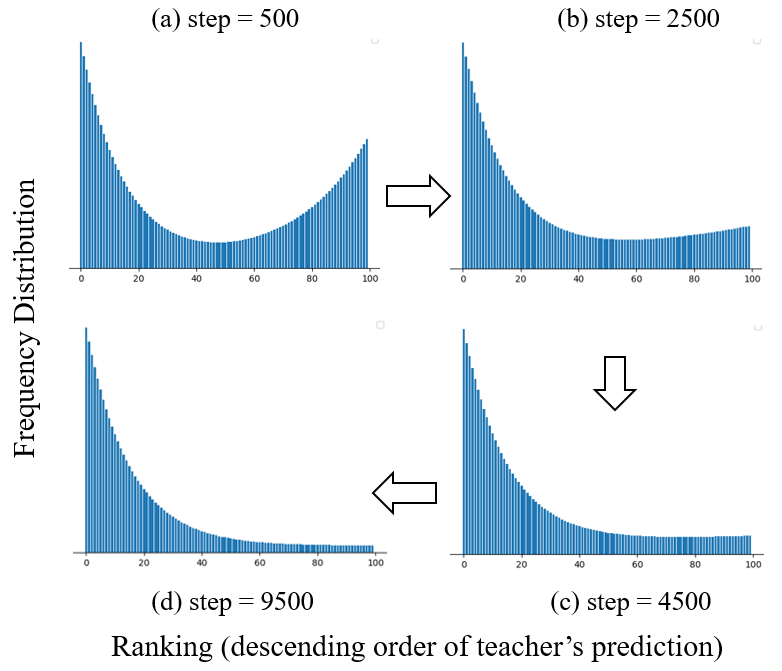}
\caption{Frequency distribution of each rank being selected during the training with PDBS method. The variation in the distribution is dataset-independent.}
\label{fig:selection_variation}  
\end{figure}

\subsection{Dynamic Mixture of Laplace Distributions}

In order to better approximate the deterministic selection process of PDBS with a stochastic distribution, we first count the frequency of each \textbf{rank} being selected during training.  Fig. \ref{fig:semantic_sorted} illustrates the difference between class and rank. By organizing the classes in descending order of the teacher's prediction, we observe some interesting patterns in the frequency statistics. We make three important observations from  Fig. \ref{fig:selection_variation}:

(1) At the early stage of training, PDBS method prefers to select classes with both high and low ranks, which corresponds to the ends in the frequency distribution. The left tip is often twice as high as the one on the right.

(2) By the middle of training, the height of the tip on the right gradually decreases.

(3) In the end, it converges to an exponential distribution and ends up selecting high rank classes more often, which forms a distribution just like the teacher's prediction over ranked classes.

\begin{figure}[t]
\centering 
\includegraphics[scale=0.38]{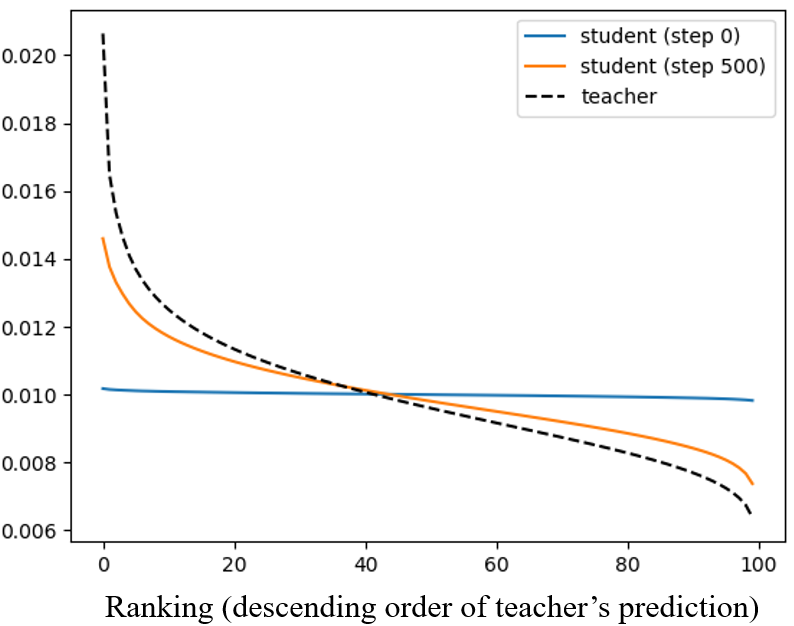}
\caption{Predictions of both the teacher and student at different training stages. At the beginning of training, the major difference between the teacher's and student's predictions is mainly distributed at the both of the ranking.}  
\label{fig:tea_stu_preds}  
\end{figure}

Moreover, we find that this pattern seems to be dataset-independent as it emerges across different datasets. The main reason of forming such peculiar distribution is because the initialization strategy we choose for the student network.  Fig. \ref{fig:tea_stu_preds} presents both the teacher's and student's predictions over a set of ranked classes. Since we initialize the weights and biases with really small floating numbers, at the beginning of the training, the prediction is almost a uniform distribution. Also, since the teacher is a trained network, its prediction is likely to have a wider range. Therefore, the prediction difference on both high-rank classes and low-rank classes are relatively large. 

\begin{figure}[t]
\centering 
\includegraphics[scale=0.35]{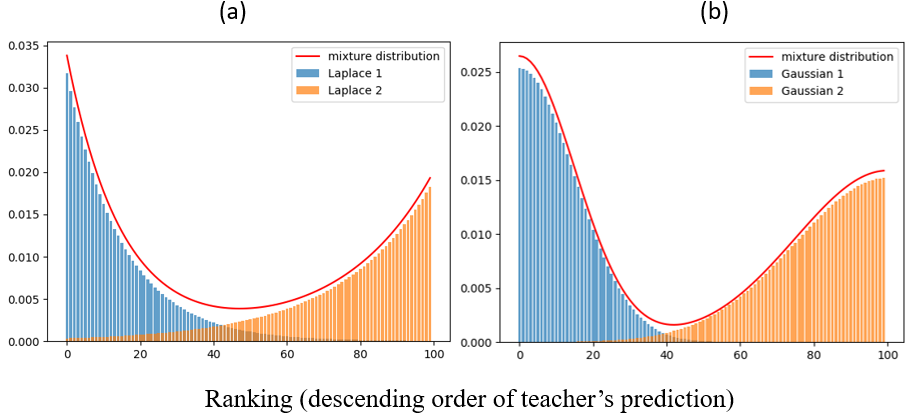}
\caption{Fitting the normalized frequency distribution over ranks in  Fig. \ref{fig:selection_variation} obtained from PDBS with (a) a mixture of two Laplace distributions (b) a mixture of two Gaussian distributions.}  
\label{fig:laplacian}  
\end{figure}

 
Based on these observation, we propose to fit the normalized frequency distribution with a mixture of two Laplace distributions as shown in  Fig. \ref{fig:laplacian} (a). In fact, we can choose any appropriate distribution to fit the frequency distribution, such as a mixture of Gaussian distributions. The reasons we choose a mixture of two Laplace distributions are: (1) Both ends in Fig. \ref{fig:selection_variation} are pointy, which are similar to the one in the Laplacian. (2) The distribution seems to decrease exponentially from the ends towards the middle. (3) It simulates the variation in Fig. \ref{fig:selection_variation} easily by increasing the scale of the second Laplace as training proceeds. 

It is also reasonable to choose other distributions like a mixture of Gaussians. However, we can see from  Fig. \ref{fig:laplacian}(b) that a Gaussian has a flatter top which doesn't approximate the frequency distribution very well. We also verify this in practice that using the mixture of Laplace distributions is slightly better than using a mixture of Gaussians.

A typical Laplace distribution is defined as follows:
\begin{equation}
    f(x) = \frac{1}{2b}exp({-\frac{\left | x-\mu\right |}{b}})
\end{equation}
where $\mu$ is the location parameter and $b$ is the scale paramter, which corresponds to the mean and variance in a gaussian distribution. To fit the frequency distribution in  Fig. \ref{fig:selection_variation}, we set $(\mu_1, b_1)=(0,3)$ for the left Laplace and set $(\mu_2, b_2)=(1,5)$ for the right one. Then we discretize the composite distribution within [0,1] into $m$ bins. Finally we normalize the mixture over all the bins. In order to further simulate the dynamics of the PDBS selection process, we fix the scale $b_1$ of the left Laplacian and linearly increase the scale $b_2$ of the second. In this way, the right Laplacian will gradually converge to a uniform distribution, which makes the overall distribution similar to the one in  Fig. \ref{fig:selection_variation}(d). 

We stress again that this mixture distribution is defined over a set of ranks. During training, we sample a subset of ranks for each mini-batch and then find the corresponding weight vector of each rank as shown in  Fig. \ref{fig:ours}. Following this method, we obtain an effective approximation to the PBDS selection process without computing the full softmax. Combined with the importance-sampling based distillation, we present the dynamic importance sampling (DIS) method to accelerate large scale distillation, which reduces the computational costs significantly while maintaining competitive performance. For comparison, we also develop a method called \textit{fixed-teacher-importance-sampling} (FTIS) which uses the prediction of the teacher as the proposal distribution. Experiments show that our approach beats the FTIS method as well as other sampling based methods on benchmark datasets.

\section{Experiments}
We compare our approach with distillation and other baseline methods on two benchmark datasets. We adopt the notations in \cite{ba2014deep} to denote the model structures. We evaluate the model performance using different metrics, as well as compare the computational costs of each method with different hyperparameters.

\subsection{Datasets}
Experiments are conducted on two datasets. The CIFAR-100 \cite{krizhevsky2009learning} consists of 60,000 32x32 color images in 100 classes, and we use 50,000 samples as the training set and the rest for testing. Market-1501 \cite{zheng2015scalable} is a benchmark dataset in person re-identification problem that requires algorithms to spot a person of interest across different camera views. The dataset consists of 32,668 images of 1,501 identities captured from 6 non-overlapping camera views. We use 751 identities for training and the rest for testing.

\begin{table}[t]
  \centering
  \begin{tabular}{l@{\hskip .6in}c@{\hskip 0.2in}c}
  \toprule
    Methods  & Accuracy \\\midrule
    teacher (ResNet32)  &  69.42\%\\
    student (shallow CNN) & 37.39\% \\
    distillation &44.28\% \\ \hline
    PDBS (k=10) &44.61\% \\
    uniform (k=10) &42.79\% \\
    FTIS (k=10) &\textbf{44.84\%} \\
    DIS (k=10) &\textbf{45.16\%} \\\bottomrule
  \end{tabular}
  \caption{Top-1 classification accuracy of the shallow CNN trained by various methods on CIFAR-100 dataset. We pick the optimal hyperparameters for each method using grid search.}
  \label{tbl:cifar100_scn}
\end{table}

\begin{table}[t]
  \centering
  \begin{tabular}{l@{\hskip .7in}c@{\hskip 0.2in}c}
\toprule
    Methods  & Accuracy \\\midrule
    teacher (ResNet32)  &  69.42\%\\
    student (LeNet) & 39.63\% \\
    distillation &46.35\% \\ \hline
    PDBS (k=10) &\textbf{46.90\%} \\
    uniform (k=10) &46.27\% \\
    FTIS (k=10) &46.48\% \\
    DIS (k=10) &\textbf{47.30\%} \\\bottomrule
  \end{tabular}
  \caption{Top-1 classification accuracy of LeNet trained by various methods on CIFAR-100 dataset. We pick the optimal hyperparameters for each method using grid search.}
  \label{tbl:cifar100_lenet}
\end{table}

\subsection{Metrics}
We report \textit{top-1 classification accuracy} on CIFAR-100 dataset. As for Market-1501, we report one extra metric in information retrieval called \textit{mean average precision} (meanAP) which computes the mean of the average precision scores for each query. As for evaluating the computational costs during training, we report the runtime of computing the last softmax activation and the gradients against the corresponding top-1 classification accuracy.

\subsection{Implementation Details}
On CIFAR-100, we choose ResNet32 \cite{he2016deep} as the teacher model. For the student models, we choose LeNet \cite{lecun1998gradient} and a shallow neural network that has 1 convolutional layer with 32 5x5 kernels (stride=2) followed by a 2x2 maxpooling layer. To reduce the network parameters, We insert a 1200-dim linear bottleneck layer between the pooling layer and the last 2048 FC layer with ReLU activation. We use ADAM for Optimizer (initial learning rate=0.01, $\beta_1$=0.9, $\beta_2$=0.99) to train all the student models for 30 epochs.

On Market-1501, we use ResNet152 as the teacher model and ResNet18 as the student model. We run all the experiments run for 180 epochs. We train the student model with the original one-hot labels with dropout. Each model is trained by RMSProp optimizer (initial learning rate=0.01, momentum=0.9) for 180 epochs. We normalize the samples without performing any other data augmentation. We use the teacher model to relabel the datasets before training the students.

\subsection{Methods}
\begin{figure}[]
\centering 
\includegraphics[scale=0.35]{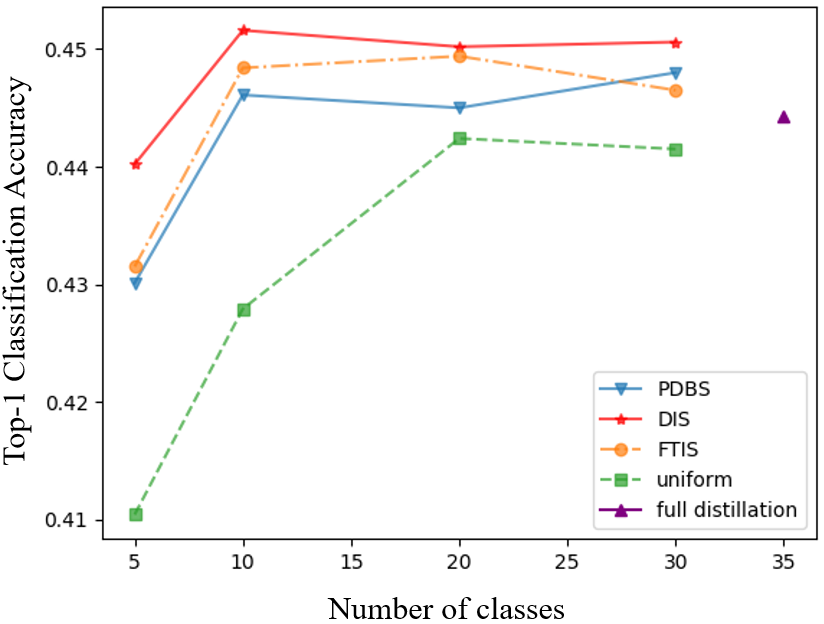}
\caption{Top-1 accuracy of the shallow CNN trained by different methods vs. the size of the selected subset on CIFAR-100 dataset. The number of classes for full distillation is 100, but for visualization purpose we set it to 35.}
\label{fig:classes_vs_acc}  
\end{figure}

We train each student model by all the approaches mentioned below with different hyperparameters:

\noindent\textbf{(1) Hard Labels: } A regular training method where the student is trained with the one-hot labels.

\noindent\textbf{(2) Knowledge Distillation: } A model transfer technique that enables a shallow student model to learn from a well-performing teacher by minimizing the cross entropy between the two predictions.

\noindent\textbf{(3) Prediction-Difference based Selection (PDBS): } A heuristic class selection approach which selects classes that have the biggest prediction difference between the teacher and student and then compute the partial softmax over the selected subset. However, this method needs to compute the full softmax activation to obtain the prediction difference.

\noindent\textbf{(4) Uniform Sampling: } An importance-sampling based distillation that uses the uniform distribution as the proposal distribution for each sample.

\noindent\textbf{(5) Fixed Teacher Importance Sampling (FTIS): } An importance-sampling based distillation that uses the teacher's prediction as the proposal distribution for each corresponding sample.

\noindent\textbf{(5) Dynamic Importance Sampling (DIS): } An importance-sampling based approach that uses the mixture of Laplace distributions as the proposal distribution for each mini-batch. Note that we can only sample ranks from this distribution and we need to convert those ranks to the corresponding classes as shown in  Fig. \ref{fig:ours}. The mixture of distributions varies while training.

\subsection{Results on CIFAR-100}
Tables \ref{tbl:cifar100_scn} and \ref{tbl:cifar100_lenet} summarize the results on CIFAR-100 dataset. We select the optimal hyperparameters for each method given the size of the selected subset.  Fig. \ref{fig:classes_vs_acc} illustrates the trade-off between the number of classes and the performance of each method. We can conclude from these results that: (1) All the sampling based or selection based approaches reach the similar accuracy when the size of the subset is large. (2) The adaptive guided sampling method has the highest performance and is the most stable one over different sizes of the subset. (3) The FTIS method achieves similar performance to the PDBS method, but it is still worse than our propsed DIS method. (4) uniform sampling performs the worst among all those methods. However, it still manages to surpass the one trained by the original one-hot labels, suggesting that even a little knowledge from the teacher can be very helpful.

The performance of either sampling from a fixed teacher or the pure selection method implicates that introducing randomness in the selection process and back-propagating the gradients with largest absolute value are equally important. Our method combines the advantages of both the sampling based approach and selection based approach. 

\begin{table}[htbp]
  \centering
  \begin{tabular}{l@{\hskip .2in}c@{\hskip 0.2in}c@{\hskip 0.2in}c}
\toprule
    Methods &meanAP & Accuracy \\\midrule
    teacher (ResNet152) &63.7\%  &84.2\%\\
    student (ResNet18) &55.5\%  &79.3\% \\
    distillation &61.7\%  &82.6\% \\ \hline
    PDBS (k=20) &59.5\%  &81.9\% \\
    PDBS (k=120) &\textbf{62.0\%}  &\textbf{82.7\%} \\
    uniform (k=20) &52.8\%  &73.9\% \\
    uniform (k=120) &59.5\%  &79.8\% \\
    FTIS (k=20) &52.1\%&77.9\%\\
    FTIS (k=120) &59.3\%&\textbf{82.1}\%\\
    DIS (k=20) &58.9\% &79.6\%\\
    DIS (k=120) &\textbf{61.2\%} &81.9\%\\\bottomrule
  \end{tabular}
  \caption{Top-1 classification accuracy and meanAP of ResNet18 trained by various methods on Market-1501 dataset. We pick the optimal hyperparameter for each method using grid search.}
  \label{tbl:market1501_resnet18}
\end{table}

\subsection{Results on Market-1501}
\begin{figure}[]
\centering 
\includegraphics[scale=0.37]{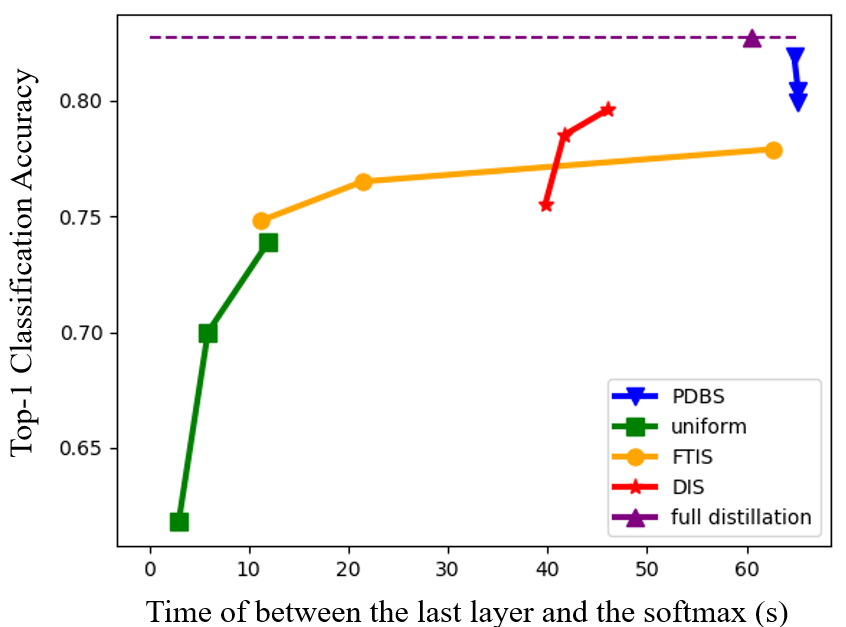}
\caption{Top-1 classification accuracy vs. computational cost of ResNet18 trained by different methods on Market-1501. Points closer to the upper left means high accuracy with low computational costs. }
\label{fig:cost_vs_acc}  
\end{figure}

 Fig. \ref{tbl:market1501_resnet18} summarizes the meanAp, allshots and the top-1 classification accuracy of the student model trained by different methods on Market-1501.  Fig. \ref{fig:cost_vs_acc} compares the computational cost against the performance for various methods. We choose top-1 accuracy to characterize the performance. We can see that our method strikes a good balance between the approximation precision and computational costs. The dynamic importance sampling method reduces the time of computing the last softmax activation per iteration from 60.68s to 46.01s on a 2 GHz Intel Core i5 CPU and a Tesla m60 GPU, speeding up by 23\%. We observe that the PDBS and DIS still outperform the FTIS and uniform sampling methods, as well as achieve a really close performance to distillation. 

Fig. \ref{fig:cost_vs_acc} summarizes results of performance against costs of different methods. The computational costs of sampling from a distribution is non-negligible because of the way we implement it.
For FTIS method which needs to sample the teacher's prediction for every sample, the run-time goes up quickly as the size of the subset increases. Those sampling methods could have used less time if we had optimized the procedure of the sampling process. However, even with non-negligible extra overheads which could be avoided, our method still reduces the training time by a large margin while maintaining a competitive performance. The experiment results on Market-1501 further prove the effectiveness of our dynamic importance sampling method.

\section{Related Work}
\subsection{Knowledge Distillation}
Model compression \cite{bucilua2006model} aims to compressing a large complex model into a smaller one without significant loss in performance. For models like neural networks, a direct approach is to minimize the L2 loss between the two networks' logits \cite{ba2014deep}. Knowledge distillation \cite{hinton2015distilling} is also one of the model compression methods which transfers the knowledge within a teacher model to a shallower student model. Other than the accuracy improvement, distilling knowledge from a deep neural net to models, like decision tree, helps to interpret how the network makes decisions \cite{frosst2017distilling}. Moreover, methods that utilize the teacher model's intermediate layers to guide the training of a student also provide extra benefits to train very deep models. \cite{romero2014fitnets,luo2008can,huang2017like}. The born-again network distills the knowledge to itself in order to learn from the past experience  \cite{furlanello2018born}. There are also works that try to provide a theoretical explanation for distillation by unifying with the privileged information theory  \cite{lopez2015unifying}. Distillation can also be applied to meta-learning such as transferring the attention map of a deep CNN \cite{zagoruyko2016paying}. 

\subsection{Approximate Softmax}
Sampling based approaches \cite{jean2014using} samples a small subset of the full classes. In hierarchical softmax \cite{morin2005hierarchical} the flat softmax layer is replaced with a hierarchical layer that has the words as leave nodes. Differentiated softmax \cite{chen2015strategies} is based on the intuition that not all the words need the same number of paramters to fit to. There are also selection based approaches \cite{zhang2018accelerated} which are designed to pick the classes according to some heuristics. Most of the approaches mentioned above mean to lower the computational cost of the softmax activation, however, in some cases the model performance was improved by using those approximation methods \cite{jean2014using,zhang2018accelerated}. It is also applicable to replace the sampling-based method in our work with any of the above approximation methods to accelerate knowledge distillation. However, we find the sampling approach is more intuitive and more compatible with distillation in our initial exploratory experiments.  

\section{Discussion}
Our experimental results show that gradients computed over a subset of classes can provide effective approximation with proper selection approaches. We have already presented two kinds of method for selecting such subsets: importance sampling and heuristic selection. The purposes of these two approaches are actually the same, which are to approximate the expected gradients of the energy function as accurately as possible. A difference between these two methods is whether to introduce randomness in the selection process. Our results in  Fig. \ref{fig:classes_vs_acc} demonstrate that introducing noises while selecting the subset sometimes can provide extra regularization on the representation. As we can see, when the number of classes are extremely small, the performance of our DIS method does not seem to drop as quickly as other methods. 

\section{Conclusions}
In this work, we present a novel importance-sampling based method which not only reduces the computational costs for large scale distillation, but also sometimes even outperforms the original distillation method. We highlight the utility of our dynamic distribution which is derived from the frequency statistics of the prediction-difference based selection. By sampling from this prior, we save the cost from querying the full softmax activation while maintaining the major information to back-progagate. Experiments on large scale datasets show that our proposed method can accelerate the training speed by a large margin without significant loss in precision.

{\small
\bibliographystyle{ieee}
\bibliography{citation}
}

\end{document}